\documentclass{bmvc2k}
\usepackage{xspace}
\usepackage{amssymb}
\usepackage{booktabs}
\usepackage{multirow}
\usepackage{amssymb}
\usepackage{pifont}
\usepackage{capt-of}
\usepackage{ragged2e}
\usepackage{ifthen}

\newcommand{\tntEnconderName}{Task-Conditioned Video Encoder}
\newcommand{\tntEnconder}{g}
\newcommand{\tntConditionerName}{Task Conditioner}
\newcommand{\tntConditioner}{\Psi}
\newcommand{\tntClassifierName}{Task-Conditioned Transductive Classifier}
\newcommand{\tntClassifier}{h}
\newcommand{\tntTaskEncoderName}{Task Encoder}
\newcommand{\actionlabel}{a}
\newcommand{\textdescription}{x}
\newcommand{\video}{v}
\newcommand{\videoAfterEnconder}{\textbf{v}}
\newcommand{\setVideosAfterEnconder}{\textbf{V}}
\newcommand{\supportset}{\mathcal{S}}
\newcommand{\queryset}{\mathcal{Q}}
\newcommand{\tntfunction}{f}

\newcommand{\cmark}{\ding{51}}%
\newcommand{\xmark}{\ding{55}}%
\RequirePackage{color}

\newboolean{main_document}

\title{TNT: Text-Conditioned Network with Transductive Inference for Few-Shot Video Classification}

\addauthor{Andrés Villa}{afvilla@uc.cl}{1}
\addauthor{Juan-Manuel Perez-Rua$^\ast$}{jmpr@fb.com}{2}
\addauthor{Victor Escorcia$^\dagger$}{v.castillo@samsung.com}{2}
\addauthor{Vladimir Araujo}{vgaraujo@uc.cl}{1,4}
\addauthor{Juan Carlos Niebles}{jniebles@cs.stanford.edu}{3}
\addauthor{Alvaro Soto$^\dagger$}{asoto@ing.puc.cl}{1}

\addinstitution{
 Pontificia Universidad Católica de Chile\\
 Santiago, Chile
}
\addinstitution{
 Samsung AI Centre Cambridge\\
 Cambridge, UK
}
\addinstitution{
 Stanford University\\
 Stanford, CA, USA
}
\addinstitution{
 KU Leuven\\
 Leuven, Belgium
}

\runninghead{VILLA, PEREZ-RUA, ESCORCIA, ARAUJO, NIEBLES, SOTO}{TNT}

\def\eg{\emph{e.g}\bmvaOneDot}

\begin{document}

\maketitle
\renewcommand\thefootnote{$\dagger$}
\footnotetext{Equal advising.}

\renewcommand\thefootnote{$\ast$}
\footnotetext{Current affiliation: Facebook AI, London, UK.}

\begin{abstract}

Recently, few-shot video classification has received an increasing interest. Current approaches mostly focus on effectively exploiting the temporal dimension in videos to improve learning under low data regimes. However, most works have largely ignored that videos are often accompanied by rich textual descriptions that can also be an essential source of information to handle few-shot recognition cases. In this paper, we propose to leverage these human-provided textual descriptions as privileged information when training a few-shot video classification model. Specifically, we formulate a text-based task conditioner to adapt video features to the few-shot learning task. 
Furthermore, our model follows a transductive setting to improve the task-adaptation ability of the model by using the support textual descriptions and query instances to update a set of class prototypes.
Our model achieves state-of-the-art performance on four challenging benchmarks commonly used to evaluate few-shot video action classification models.
\end{abstract}

\section{Introduction}

Humans use language to guide their learning process~\cite{HLPM}. For instance, when teaching how to prepare a cooking recipe, visual samples are often accompanied by detailed or rich language-based instructions~(\eg, ``\textit{Place aubergine onto pan}"), which are fine-grained and correlated with the visual content. 
\begin{figure}[ht]
\centering
\includegraphics[width=1.0\linewidth]{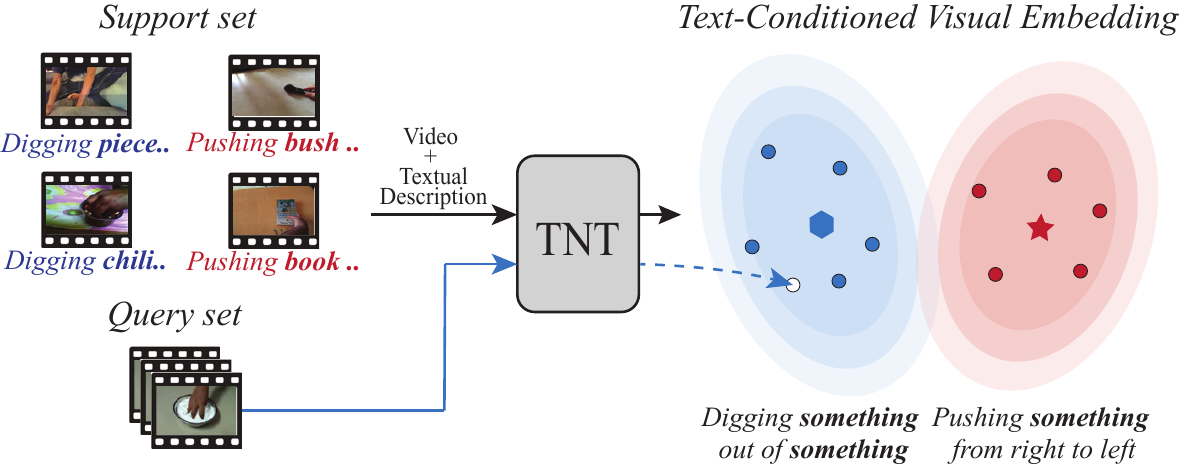}
\vspace{-10pt}
\caption{{\bf Outline of our FSL setting.} Our model leverages the rich text descriptions of the support instances (left) to improve class discrimination (right) in two different ways.
1) Modulating the visual feature encoder to alleviate the large intra-class variations of video data. 2) A transductive setting where textual information of the support instances is used alongside visual information of the query set to augment the support set.}
\vspace{-4mm}
\label{fig:teaser_tnt}
\end{figure}
These instructions are a primary cause of the human ability to quickly learn from few examples because they help to transfer learning among tasks, disambiguate and correct error sources~\cite{HLPM}. However, modern deep learning approaches in action recognition~\cite{Jingwei:EtAl:2018,DBLP:TSM, TSN} have mainly focused on a large amount of labeled visual data ignoring the textual descriptions that are usually included along with the videos  \cite{Girdhar2019VideoAT, kinetics}. 
These limitations have motivated an increasing interest in Few-Shot Learning (FSL) \cite{FSLSurvey}, which consists of learning novel concepts from few labeled instances. 




While most FSL models are focused on image classification \cite{liu2020prototype,BateniSimpleCNAPS,kim2019edgelabeling,cnapsRequeima,liu2018learning,pmlr-v70-finn17a,rajeswaran2019meta,protoNet,8578229,vinyals2016matching}, few works \cite{Cao_2020_TAM,9022182,mishra2018generative,10.1145/3206025.3206028,zhang2020fewshot,compoundMem,zhu2021few} are dedicated to video classification. Recognizing actions in video with only a few training samples is arguably more challenging than the image classification case. 
The video content is richer, and action classes exhibit large intra-class variations. For example, the action "Digging something out of something" in Fig.~\ref{fig:teaser_tnt} looks significantly different as it involves interactions with two different objects. Therefore, extending existing FSL approaches for image classification to the case of video is not trivial.

The few existing video FSL methods follow one of two approaches: (i) exploiting the temporal and spatial dimensions in videos \cite{Cao_2020_TAM,zhang2020fewshot}; or (ii) taking advantage of large amounts of additional video data by using tag retrieval to overcome the labeled data scarcity \cite{Xian2020GeneralizedMF}. 
However, recent work has not explicitly leveraged the available natural language descriptions that come with videos as an additional information source. These descriptions can be easily obtained without further effort while the dataset is collected, as described by \cite{epickitchens}. During the Epic-Kitchens \cite{epickitchens} collection, the actors simply narrated their actions using free-form language. 
We found that these text descriptions are crucial to recognizing actions in a few-shot regime, which agrees with the human ability to compound and exploit multimodal knowledge to learn from few training samples quickly.
In this paper, for the first time, we introduce a new class of models: \textbf{T}ext-conditioned \textbf{N}etworks with \textbf{T}ransductive inference or TNT. This method exploits the knowledge that is available in text descriptions as a privileged source of information~\cite{vapnik2009new} to improve class discrimination in few-shot video classification, see Fig.~\ref{fig:teaser_tnt}. TNT is built on top of a primary backbone that aims to encode global and extensive knowledge about the visual world.
TNT further contains a complementary secondary network trained to extract task-specific knowledge from the support textual descriptions, leveraging the modern language models~\cite{reimers2019sentencebert}. This secondary network contextualizes the global knowledge of the primary network according to the semantic information of the task. Moreover, TNT spans a third module, which leverages the detailed textual information of the few support videos to augment them with those unlabeled (query) to obtain more confident class representations (prototypes), following a transductive setup. These prototypes serve as a proxy for the classification of the query instances using a nearest neighbors approach. Overall, the integration of these three networks allows our model to quickly adapt to the challenging data conditions of FSL tasks.

In summary, our main contributions are: ({\bf I}) To the best of our knowledge, we propose the first FSL video action classification method that leverages the semantic information in textual action descriptions of the support data to modulate the visual feature encoder. ({\bf II}) We show the advantage of using the semantic information in support textual action descriptions to perform transductive learning. We develop a dynamic prototype module that uses textual semantic representations to obtain class prototypes using both labeled and unlabeled samples following an attentive approach. ({\bf III}) We demonstrate that textual embeddings outperform the video ones for task adaptation even when these descriptions are short and class-specific (\eg, class labels: \textit{Headbanging}, \textit{Stretching leg}, etc). ({\bf IV}) We achieve state-of-the-art performance with two families of video action FSL benchmarks, those with detailed or rich textual descriptions such as Something-Something-100 (SS-100) \cite{Cao_2020_TAM} and the new benchmark Epic-Kitchens-92 (EK-92), and those with short class-level textual descriptions such as MetaUCF-101~\cite{mishra2018generative} and Kinetics-100~\cite{compoundMem}. 
\vspace{-3mm}

\section{Related Work}
\noindent\textbf{Few-Shot Learning.}
It is possible to identify two main groups in the FSL literature: 
(i) gradient based methods and (ii) metric learning based methods. 
Gradient-based methods focus on learning a good parameter initialization that facilitates model adaptation by few-shot fine-tuning \cite{pmlr-v70-finn17a,nichol2018firstorder,rajeswaran2019meta}. On the other hand, metric-based methods aim to learn or design better metrics for determining similarity of input samples in the semantic embedding space~\cite{koch2015siamese, qi2018low, protoNet, 8578229, vinyals2016matching}.
More recently, affine conditional layers are added to the feature extraction backbone in~\cite{BateniSimpleCNAPS,cnapsRequeima} as extension to the conditional neural process framework~\cite{pmlr-v80-garnelo18a} with the goal of effective task-adaptation.
In this work, we extend this framework~\cite{pmlr-v80-garnelo18a} differently from~\cite{BateniSimpleCNAPS,cnapsRequeima} by adapting the feature extractor and updating the class representations based on the support textual descriptions and query instances. Our goal is to influence the visual backbone with the structured knowledge captured by pre-trained language models.

\noindent\textbf{Induction vs Transduction in FSL.}
Regarding the inference setup, there are two types of approaches: inductive and transductive FSL. In the inductive setting, only the support instances are used to guide the inference process. In contrast, in the transductive setting, the model uses extra information from query samples to perform its inference \cite{liu2020prototype}. 
We are motivated by recent work following the transductive setting \cite{kim2019edgelabeling,liu2020prototype,liu2018learning,nichol2018firstorder}, where the unlabeled query data is exploited to further refine the few-shot classifier. For instance, \cite{liu2020prototype} proposes a prototype rectification approach by label propagation. %
Departing from previous work, our model proposes a novel transductive approach that takes advantage of the support textual descriptions to augment the support videos with the unlabeled instances, leveraging the cross-attention approach. 

\noindent\textbf{Few-Shot Video Classification.}
With the shift of action recognition research from
coarse~\cite{kinetics} to fine-grained categories~\cite{epickitchens,S2Sv2Goyal}, the problem of data scarcity has intensified. A few works to tackle this issue have appeared recently. However, most of works focused only on better exploiting visual or temporal information from videos \cite{Cao_2020_TAM,9022182,mishra2018generative,Xian2020GeneralizedMF,10.1145/3206025.3206028, zhang2020fewshot,compoundMem,zhu2021few}. Additionally, the approach proposed in \cite{Xian2020GeneralizedMF} uses extra video data and annotations to learn a more suitable representation before meta-training.
Although tackling important aspects in video data modeling, none of the previous works offer solutions to the semantic gap between the few-shot samples and the nuanced and complex concepts needed for video representation learning. We aim to bridge this gap by using textual descriptions as privileged information to contextualize the video feature encoder in conjunction with a classification approach based on class prototypes acting under a transductive inference scheme.

\noindent\textbf{Exploiting Text Embeddings in Small Data.}
Prior works have leveraged multi-modal information to enhance few-shot visual classification \cite{9102883,8451372,NIPS2019_8731,10.1145/3206025.3206028}, where textual description has been widely used for image data. Likewise, Zero-Shot Learning (ZSL) methods for image classification uses text descriptions to classify samples from novel unseen classes~\cite{lee2018multi,NIPS2013_5027}. These descriptions focus on nouns that have a structured taxonomy and can be associated with specific regions of the input image. Conversely, actions are defined by verbs that are usually more overloaded and more fine-grained than nouns \cite{caba2015activitynet}. Therefore, the extrapolation of these approaches to the case of action recognition is not straightforward.
Currently, there are some relevant works in ZSL for video classification~\cite{Brattoli_2020_CVPR,action2vecHahn,Jain_2015_ICCV, GaoEtAl, video2vec}. \cite{Brattoli_2020_CVPR,action2vecHahn,video2vec} learn a static video encoder to map the videos to an embedding space very close to the semantic representation of their labels. It does not allow these methods to learn to exploit the Spatio-temporal information of the videos specifically, limiting their generalization power in scenarios where some support instances are available. That is why we employ a two steps training process and adaptative method. First, we learn a general video encoder from base classes. Later, the general video encoder is fixed, and our method learns to adapt the video encoder and a transductive classifier to the novel classes using textual descriptions. Likewise, \cite{Jain_2015_ICCV, GaoEtAl} use a static video representation to get the relevant objects in the video and later computes their semantic textual embedding to be the bridge between known and unknown actions labels. In this sense, this method depends on the semantic relation between the action label and its objects, which could be a problem for fine-grained action datasets like \cite{S2Sv2Goyal}, where the objects are related to several classes of actions.

\section{Method}
\vspace{-2mm}

\subsection{Problem Definition}
FSL aims to obtain a model that can generalize well to novel classes with few support instances. Therefore, we follow the standard FSL setting~\cite{protoNet,vinyals2016matching}, wherein a trained model $\tntfunction_\theta$ is evaluated on a significant number of $N-$way $K-$shot tasks sampled from a meta-test set $D_{test}$. These tasks consist of $N$ novel categories, from which $K$ samples are sampled to form support set $\supportset$, where $K$ is a small integer, typically, 1 or 5. The support set $\supportset$ is used as a proxy to classify the $B$ unlabeled instances from the query set $\queryset$. The parameters $\theta$ of the model $f$ are trained on a meta-training set $D_{train}$, by applying the episodic training strategy proposed by~\cite{vinyals2016matching}. This is, $N-$way $K-$shot classification tasks are simulated by sampling from $D_{train}$ during meta-training. $\queryset$ is sampled from the same $N$ categories in such a way that the samples in $\queryset$ are non-overlapping with $\supportset$.
The set of classes available for meta-training are often referred to as base classes. Note that the model $\tntfunction$ is evaluated on different categories than it is trained on. In this paper, we assume that a text description is available for each instance in $\supportset$.

\begin{figure*}[ht]

\centering
\includegraphics[width=1.0\linewidth]{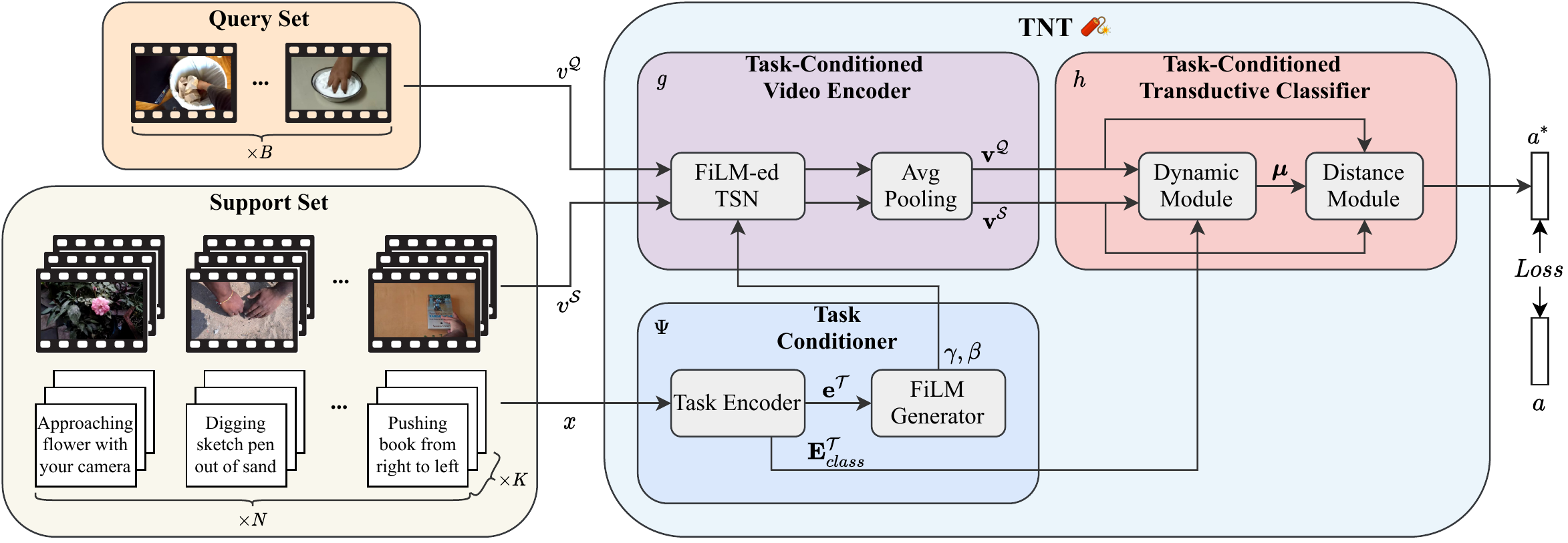}
\vspace{-7mm}
\caption{
TNT model is composed by three parts. 
(I) \tntEnconderName \space $\tntEnconder$ generates representations $\textbf{v}^\mathcal{Q}, \textbf{v}^\mathcal{S}$ of video sequences conditioned on parameters $\beta$ and $\gamma$. 
(II) \tntConditionerName \space $\tntConditioner$ takes video descriptions $x$ to compute the text embeddings $\textbf{e}^\mathcal{T}$ for generating modulation parameters $\beta$ and $\gamma$, and the semantic class embedding $\textbf{E}^\mathcal{T}_{class}$.
(III) \tntClassifierName \space $\tntClassifier$ takes the video representations $\textbf{v}^\mathcal{Q}, \textbf{v}^\mathcal{S}$ and the embedding $\textbf{E}^\mathcal{T}_{class}$ to classify unlabeled samples following a transductive approach.
}
\vspace{-4mm}
\label{fig:main_model}
\end{figure*}

\subsection{TNT Model}
We strive for action classification in videos within a low-data setting by means of (i) the rich semantic information of textual action descriptions and (ii) exploiting the unlabeled samples at test time. We accomplish this task with our \textbf{T}ext-Conditioned \textbf{N}etworks with \textbf{T}ransductive Inference (TNT), depicted in Fig.~\ref{fig:main_model}. Our overall model $\tntfunction$ is a text-conditioned neural network designed to be flexible and adaptive to novel action labels. Taking inspiration from \cite{BateniSimpleCNAPS,cnapsRequeima}, TNT is composed by three modules: (i) \tntEnconderName \space $\tntEnconder$; (ii) \tntConditionerName \space  $\tntConditioner$; and (iii) \tntClassifierName \space $\tntClassifier$.

\noindent\textbf{\tntEnconderName.} This module $\tntEnconder$ transforms the lower-level visual information of each video $\video$ into a more compact and meaningful representation $\videoAfterEnconder$.
To handle novel action classes at test time, it is essential to provide $\tntEnconder$ with a flexible adaptation mechanism that selectively focuses and/or disregards the latent information of its internal representation across different episodes. To achieve this, we employ the TSN video architecture with a ResNet backbone that is enhanced by adding Feature-wise Linear Modulation (FiLM) layers after the BatchNorm layer of each ResNet block. FiLM layers adapt the internal representation $\videoAfterEnconder_i$ at the $i^{th}$ block of $\tntEnconder$ via an affine transformation $FiLM(\videoAfterEnconder_i; \gamma_i ,\beta_i) = \gamma_i\videoAfterEnconder_i + \beta_i$ where $\gamma_i$ and $\beta_i$ are the modulation parameters generated by the \tntConditionerName~ module.
Thereby, this module computes frame-level feature embeddings for each video followed by an adaptive average pooling that summarizes the spatiotemporal information to obtain the video representation $\videoAfterEnconder= g(\video)$ where $\video \in \mathbb{R}^{T \times H \times W}$ and $\videoAfterEnconder \in \mathbb{R}^{G}$. 

We use the widely-adopted video frame sampling strategy of temporal segment networks (TSN)~\cite{DBLP:TSM,TSN,TRNZhou, Zolfaghari_2018_ECCV}. Contrary to the CNAPS strategy \cite{BateniSimpleCNAPS,cnapsRequeima}, we train \tntEnconderName~on the base classes within a fully supervised regime rather than on a large dataset. That is due to the variability between the video datasets and their actions. So that, we have to train it for few epochs to avoid overfitting. 


\noindent\textbf{\tntConditionerName.} The \tntConditionerName~$\tntConditioner$ is an essential part of our approach that provides high adaptability to our model. Specifically, it computes conditioning signals that modulate the \tntEnconderName~$\tntEnconder$~and the \tntClassifierName~$\tntClassifier$~based on the textual action descriptions of a set of support instances $\supportset$.
Due to the inherent semantically rich and structured nature of textual action descriptions, we argue that explicitly exploiting text embeddings associated with action labels is crucial to adapt our model on each episode. 

\begin{figure}
\centering
\begin{tabular}{cc}
\bmvaHangBox{
\includegraphics[scale=0.5]{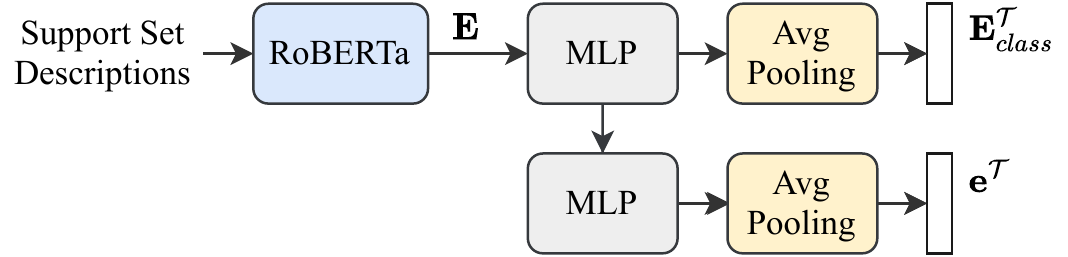}
\label{fig:text_encoder}
} & \multirow{4}{7cm}{
\centering
\bmvaHangBox{\includegraphics[scale=0.6]{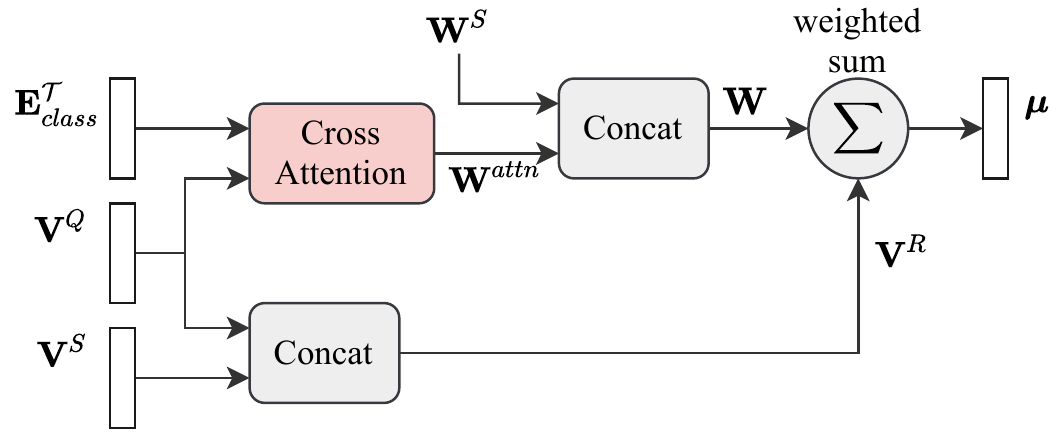}
\label{fig:dynamic_mod}}\\(c)
} \\(a) & \\
\bmvaHangBox{\includegraphics[scale=0.5]{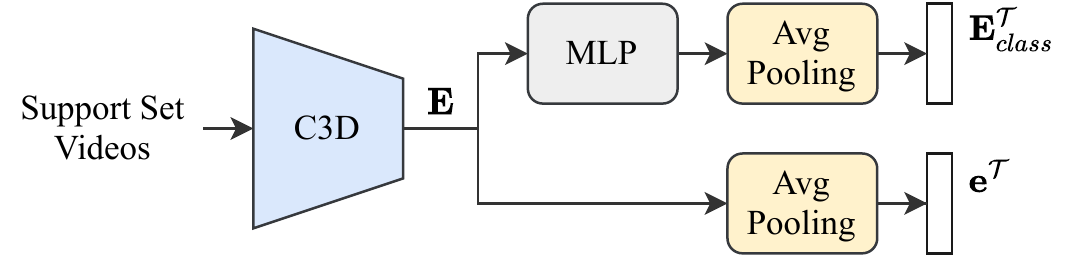}
\label{fig:video_encoder}} & \\(b) &  
\end{tabular}
\caption{
\textbf{Task conditioner architectures and Dynamic prototype module.} We propose two kinds of task conditioner: (a) a text-based conditioner based on the RoBERTa model. (b) a video-based conditioner, and (c) a dynamic prototype module based on an attention approach to augment the support set $\supportset$ samples. 
}
\vspace{-4mm}
\label{fig:encoders_dynaMod}
\end{figure}

We assume that the instances of the support set $\supportset$ are a triad $(\video,  \textdescription, \actionlabel)$ corresponding to a video \video, textual action description \textdescription, and categorical action label \actionlabel, respectively. Furthermore, the \tntConditionerName~subsumes two components:

\textbf{(a) \tntTaskEncoderName.} This module generates the conditioning signals: (i) the task embedding $\textbf{e}^{\mathcal{T}}$ to tune the \tntEnconderName~\tntEnconder, and (ii) the semantic class embedding $\textbf{E}_\mathtt{class}^{\mathcal{T}}$ used to tune the \tntClassifierName, given the textual action description $\textdescription$ in the support set $\supportset$.
Specifically, our \tntTaskEncoderName~consists of the RoBERTa language model~\cite{reimers2019sentencebert} followed by two multilayer perceptrons, as shown in Fig.~\ref{fig:encoders_dynaMod}-a. Using RoBERTa, we compute the sample-level text embedding $\textbf{E}$ of each $\textdescription$. 
These text representations are projected first through linear layer and average-pooled along the number of shots $K$, resulting in the class embedding $\textbf{E}_\mathtt{class}^{\mathcal{T}} \in \mathbb{R}^{N \times G}$. Additionally, $\textbf{E}$ is linearly projected a second time to obtain the task embedding $\textbf{e}^{\mathcal{T}} \in \mathbb{R}^{1 \times L}$.

\textbf{(b) FiLM Generator.} It generates the set of affine parameters ${\gamma_i,\beta_i}$ for every stage $i$ of $\tntEnconder$ to effectively modulate our \tntEnconderName~given the task embedding $\textbf{e}^{\mathcal{T}}$.

In practice, we tune the MLP modules and the FiLM generator parameters in a subsequent training stage after fixing~$\tntEnconder$ \cite{BateniSimpleCNAPS}. The RoBERTa module is initialized from a pre-trained sentence representation and remained unchanged to take advantage of its prior knowledge, avoiding overfitting due to its high number of parameters. Also, note that our \tntConditionerName~module is conceptually different to the one presented in~\cite{BateniSimpleCNAPS,cnapsRequeima}. While the encoder in~\cite{BateniSimpleCNAPS,cnapsRequeima} is purely a function of the visual instances in the support set $\supportset$, in our case, we leverage textual descriptions of target categories.


 


\noindent\textbf{\tntClassifierName.} This module $\tntClassifier$ follows a metric learning approach to classify the unlabeled samples of $\queryset$ by matching them to the nearest class prototype. To obtain the class prototypes, a straightforward approach is to compute a class-wise average by considering the $K$-examples in the support set $\supportset$ \cite{BateniSimpleCNAPS, chen2020closer,protoNet}. However, due to the data scarcity, these prototypes are usually biased. To alleviate this problem, we use a transductive classifier that leverages the unlabeled samples to improve the class prototypes based on the semantic class embedding $\textbf{E}_\mathtt{class}^{\mathcal{T}}$. Specifically, the \tntClassifierName~consists of two components:

\textbf{(a) Dynamic Prototype Module.} This module leverages the semantic class embedding $\textbf{E}_\mathtt{class}^{\mathcal{T}}$ to get the most relevant unlabeled samples for every class, see Fig.~\ref{fig:encoders_dynaMod}-c. Thus, effectively augmenting the support set with unlabeled samples in $\queryset$ and subsequently improving the class prototypes. Specifically, we employ a cross-attention layer~\cite{att_all_you_need} 
to compute class-dependent relevance weights for each of the $B$ samples in the query set $\textbf{W}^{att} \in \mathbb{R}^{N \times B}$ by:

\begin{equation}
\textbf{W}^{att} = \mathtt{softmax}\left(\frac{\textbf{E}_\mathtt{class}^{\mathcal{T}}\textbf{W}^Q \left(\setVideosAfterEnconder^\mathcal{Q}\textbf{W}^K\right)^T}{\sqrt{G}}\right),
\end{equation}
where $\textbf{W}^Q$, $\textbf{W}^K$ $\in \mathbb{R}^{G \times G}$ are linear projections for the query and keys. 
In our case, we use $\textbf{E}_\mathtt{class}^{\mathcal{T}}$ as query to look up into the sequence of all video representations $\setVideosAfterEnconder^\mathcal{Q}=[\textbf{v}^{\mathcal{Q}}_1, ..., \textbf{v}^{\mathcal{Q}}_B]$ in $\queryset$. Furthermore, we calculate relevance weights $\textbf{W}^\mathcal{S} \in \mathbb{R}^{N \times NK}$ for the support samples in $\mathcal{S}$, we assume that all of them have equal importance for the class $i$ they belong to.
Thus, we define the relevance weights $\textbf{W} = [\textbf{W}^{att}~~ \textbf{W}^\mathcal{S}]$ for all the task samples $\mathcal{R} = \mathcal{Q} \cup  \mathcal{S}$, as
\begin{equation}
\textbf{W}_{ij} = \Bigg\{ \begin{matrix}
\textbf{W}^{att}_{ij}, & \videoAfterEnconder^{\mathcal{R}}_j \in \mathcal{Q}\\ 1/K, & i = a_j, \left(\videoAfterEnconder^{\mathcal{R}}_j, a_j\right) \in \mathcal{S} \\ 0, & i \neq a_j, \left(\videoAfterEnconder^{\mathcal{R}}_j, a_j\right) \in \mathcal{S}
\end{matrix},
\end{equation}

 \noindent where $\textbf{W} \in \mathbb{R}^{N \times (B+ NK)}$. Finally, we calculate the class prototypes through a weighted sum of all samples $\textbf{V}^\mathcal{R}$ attending the relevance weights $\textbf{W}$:
\begin{equation}
\boldsymbol{\mu}_{i} = \frac{1}{\sum_{j=1}^{B+NK}\textbf{W}_{ij}} \sum_{j=1}^{B+NK}\textbf{W}_{ij}\videoAfterEnconder^\mathcal{R}_j,  i \in N
\label{eq:weighted_sum}
\end{equation}


\textbf{(b) Distance Module.} 
This module classifies the unlabeled instances of the query set by matching them to the nearest class prototype. To compute the distance between each instance and prototypes, we use a class-covariance-based distance (Mahalanobis) as in \cite{BateniSimpleCNAPS}. We train our model by minimizing: 
$p(a_j^{*} = i | f(v_j), \mathcal{S}) = \mathtt{softmax}(-d_i(f(v_j^*), \boldsymbol{\mu}_i))$,
where $j \in B$, $a_j^*$ is the predicted class for the unlabeled sample $v_j^*$, and $\boldsymbol{\mu}_i$ is the prototype of class $i$ obtained with the dynamic module.
Also, $d_i$ is the distance function that receives the class prototypes explicitly and computes the task prototype by taking the average of these prototypes.

\section{Experiments}

\noindent\textbf{Datasets.}
We evaluate our approach using two families of datasets: (i) those with rich and detailed textual descriptions of actions per video: Epic-Kitchens~\cite{epickitchens}, Something-Something-V2~\cite{S2Sv2Goyal}, and (ii) those with short class-level descriptions: UCF-101~\cite{DBLP:journals/corr/abs-1212-0402} and Kinetics~\cite{kinetics}. We propose for the first time to use Epic-Kitchens~\cite{epickitchens} as a benchmark for few-shot video classification. We coin this new benchmark \textbf{Epic-Kitchens-92} (EK-92). \cite{epickitchens} features spontaneous actions accompanied with human narrations. Interestingly, a particular action class could encompass diverse narrations,~\eg, the action class: ``\textit{\underline{Put} something}" features narrations such as: ``\textit{\underline{Put}
plate \underline{down}}", ``\textit{\underline{Place} aubergine \underline{onto} pan}". To ensure that action classes are consistent, we use the 97 verb classes defined by~\cite{epickitchens} and select those with more than 5 instances, yielding 92 action classes. Then, we divide the resulting 92 classes into 58, 11, and 23 for meta-training, meta-validation, and meta-testing, respectively.
For the other benchmarks, we follow the evaluation protocol proposed by \cite{Cao_2020_TAM, mishra2018generative,compoundMem} termed  \textbf{Something-Something-100} (SS-100), \textbf{MetaUCF-101}, and \textbf{Kinetics-100}, respectively.
The protocols in \cite{Cao_2020_TAM,compoundMem} define a set of 64 classes for meta-training, 12 classes for meta-validation, and 24 classes for meta-testing, which are sampled randomly from Something-Something-V2 and Kinetics, respectively. In terms of the protocol in \cite{mishra2018generative}, it samples randomly 70 classes from UCF-101 for meta-training, 10 classes for meta-validation, and 21 for meta-testing. To facilitate the comparison, we use the same partitions proposed by the original authors.

Additionally, we make use of the provided text-based action descriptions from every meta-dataset. 
In Kinetics-100 and MetaUCF-101, we directly employ the class labels (\eg, \textit{Headbanging}, \textit{Stretching leg}, etc), which are the same for all samples that belong to the same class. Alternatively, EK-92 and SS-100 provide a fine-grained textual action description per instance based on the action and objects depicted in the video. 
Further details about these datasets can be found in the Supplementary Material.

\setboolean{main_document}{false}   
\ifthenelse{\boolean{main_document}}{
\begin{table}[t]


\centering 
\resizebox{8.3cm}{!}{%
    \begin{tabular}{cccccccc} 
    \toprule
    \multicolumn{1}{c}{\multirow{2}{*}{\bf Dataset}} &\multicolumn{3}{c}{\bf Num Classes} & \multicolumn{3}{c}{\bf Num Instances} & \multicolumn{1}{c}{\multirow{2}{*}{\begin{tabular}[c]{@{}c@{}}{\bf Rich Textual} \\ {\bf Descriptions} \end{tabular}}} \\
    \cmidrule{2-7}
         & Train & Val & Test & Train & Val & Test & \\
    \midrule
    EK-92 & 58 & 11 & 23 & 49621 & 8352 & 18370 & \cmark  
    \\
    SS-100 & 64 & 12 & 24 & 67013 & 1926 & 2857 & \cmark
    \\
    MetaUCF-101 & 70 & 10 & 21 & 9154 & 1421 & 2745 & \xmark
    \\
    Kinetics-100 & 64 & 12 & 24 & 6400 & 1200 & 2400 & \xmark
    \\
    \bottomrule
    \end{tabular}}
\caption{Statistics from the SS-100~\cite{Cao_2020_TAM}, MetaUCF-101~\cite{mishra2018generative}, and Kinetics-100~\cite{compoundMem} datasets for FSL. We introduce EK-92, a new data split from Epic-Kitchens~\cite{epickitchens}.}
\label{tab:datasets}
\end{table}
}{}

\noindent\textbf{Implementation Details.}
We train our model following the episodic learning approach to mimic the meta-testing conditions \cite{8578229}.
For this purpose, we assemble $N$-way and $K$-shot tasks, selecting $N$ classes randomly with $K$ samples for the support set and $B$ unlabeled samples for the query set. Thus, each episode has $NK+B$ videos. We report results for the 5-shot and 1-shot tasks, each with 5-ways and 50 elements of these classes in the query set (10 elements per class). 
Our model was trained during $15\times10^{3}$ episodes with the same data augmentation proposed in \cite{TSN}, using $T=8$ frames per video. 
We calculate the mean accuracy by sampling $10^{4}$ episodes (for a total of $5\times10^{5}$ queries) to test our model.
In regards to the FiLM generator and the distance module, we follow the design choices in~\cite{BateniSimpleCNAPS}. We optimize our model using task batch size of $16$ and Adam with a learning rate of $5\times 10^{-4}$ for EK-92, SS-100 and MetaUCF-101, and $1\times 10^{-4}$ for Kinetics-100.

For our feature encoder, we use a TSN Network with a ResNet-50 backbone pre-trained on ImageNet and augmented with FiLM layers. For the first training stage of the TSN backbone, we tune it during 12 epochs using the setting proposed in \cite{TSN}. 
\begin{table}[t]
\centering
\resizebox{13cm}{!}{
\bmvaHangBox{
\begin{tabular}{ccccccccc} 
    \toprule
    \multicolumn{1}{c}{\multirow{3}{*}{\bf Model}} & \multicolumn{4}{c}{\bf with Rich Textual Descriptions} & \multicolumn{4}{c}{\bf with Short Class-Level Description}\\\cmidrule(lr){2-5}\cmidrule(lr){6-9}
    & \multicolumn{2}{c}{\bf EK-92} & \multicolumn{2}{c}{\bf SS-100} & \multicolumn{2}{c}{\bf MetaUCF-101} & \multicolumn{2}{c}{\bf Kinetics-100}
    \\\cmidrule(lr){2-5}\cmidrule(lr){6-9}
         & \multicolumn{1}{c}{\bf 1-shot} & \multicolumn{1}{c}{\bf 5-shot} & \multicolumn{1}{c}{\bf 1-shot} & \multicolumn{1}{c}{\bf 5-shot} & \multicolumn{1}{c}{\bf 1-shot} & \multicolumn{1}{c}{\bf 5-shot} & \multicolumn{1}{c}{\bf 1-shot} & \multicolumn{1}{c}{\bf 5-shot} \\
    \midrule
    ARN \cite{zhang2020fewshot} & - & - & - & - & 62.1 & 84.8 & 63.7 & 82.4
    \\
    TSN++~\cite{Cao_2020_TAM} & $39.1^*$ & $52.3^*$ & $33.6$ & $43.0$ & $76.4^*$ & $88.5^*$ & 64.5 & 77.9 
    \\
    CMN++~\cite{compoundMem,Cao_2020_TAM} & - & - & 34.4 & 43.8 & - & - & 65.4 & 78.8
    \\
    TRN++~\cite{Cao_2020_TAM} & - & - & 38.6 & 48.9 & - & - & 68.4 & 82.0
    \\
    TAM~\cite{Cao_2020_TAM} & - & - & 42.8 & 52.3 & - & - & 73.0 & \textbf{85.8}
    \\
    \cmidrule{1-9} 
    TSN++ Transd. \cite{liu2020prototype} & 42.33 & 52.66 & 39.28 & 52.63 & 79.23 & 90.08 & 68.0 & 79.87 
    \\
    TNT & $\textbf{46.13} \pm 0.27$ & $\textbf{59.00} \pm 0.23$ & $\textbf{50.44} \pm 0.25$ & $\textbf{59.04} \pm 0.23$ & $\textbf{86.66} \pm 0.19$ & $\textbf{94.14} \pm 0.11$ & $\textbf{78.02} \pm 0.24$ & $84.82 \pm 0.19$ 
    \\
    \bottomrule
    \end{tabular}
} 
}
\caption{
\textbf{Results on two families of datasets}. Those with rich textual descriptions: EK-92 and SS-100. Those with class-level textual descriptions: MetaUCF-101 and Kinetics-100. We report top-1 accuracy on the meta-testing sets for the 5-way tasks. *Obtained by us.
}
\vspace{-3mm}
\label{tab:datasets_results}
\end{table}

\noindent\textbf{Baselines.} We compare the performance of our TNT model against state-of-the-art methods for few-shot video classification, namely TAM \cite{Cao_2020_TAM} and ARN \cite{zhang2020fewshot}. We also consider additional stronger baselines, namely TSN++, TRN++ and CMN++ which are proposed by \cite{Cao_2020_TAM}, following the practices from \cite{chen2020closer, compoundMem}. Because our model makes use of a transductive setting, we also consider a transductive baseline named TSN++ Transd. This baseline is an extension of the image-based method \cite{liu2020prototype} which adopts a pseudo-labeling strategy to augment the support set. Conversely, the method proposed by \cite{Xian2020GeneralizedMF} is not considered because it relies on a large amount of additional data, and its evaluation protocol is different from ours and from the one used in the baselines. Specifically, it uses the whole video instead of a segment of it.

\noindent\textbf{Impact of rich textual descriptions.} As it can be observed in Table~\ref{tab:datasets_results}, we achieve state-of-the-art-results in all standard benchmark metrics across the two tested datasets with rich textual descriptions per instance. Notably, our model achieves outstanding results in EK-92, where it must handle spontaneous and unstructured descriptions. Likewise, our model improves over the TSN++ transductive baseline by around 7\% and 4\% in the 5-shot and 1-shot tasks, respectively, which shows the relevance of using the textual descriptions to modulate the network and make a transductive inference. It is worth noting that TSN-based backbone does not have a strong temporal modeling capacity. This is in sharp contrast to TAM~\cite{Cao_2020_TAM}, which is specially designed to capture temporal information. Despite these disadvantages, our method is able to outperform this strong alternative by around 8\% in the 5-shot and 1-shot tasks on SS-100.

\noindent\textbf{Impact of short class-level textual descriptions.} Table~\ref{tab:datasets_results} shows the performance of our model in datasets without rich textual description.  Notably, our model outperforms the state-of-the-art baselines on MetaUCF-101 by a large margin. On Kinetics-100, TNT beats the TAM model, which is the best baseline in this benchmark, in the 1-shot task by 3.5\%, while in the 5-shot, the result remains competitive. A possible reason for such results is that, unlike MetaUCF-101, Kinetics-100 has significantly fewer training instances. These results suggest that TNT achieves outstanding performance with short class-level descriptions, although it is designed to leverage the rich semantic information in fine-grained textual descriptions.

\begin{table}
\centering
\resizebox{13cm}{!}{
\begin{tabular}{cc}
\bmvaHangBox{
\begin{tabular}{cccccc} 
    \toprule
    \multicolumn{1}{c}{\multirow{2}{*}{\bf Model}} &\multicolumn{1}{c}{\multirow{2}{*}{\begin{tabular}[c]{@{}c@{}}{\bf Inference} \\ {\bf Type} \end{tabular}}} &\multicolumn{2}{c}{\bf Task Encoder} & \multicolumn{2}{c}{\bf SS-100 (Val)} 
    \\\cmidrule{3-6} 
         & & \multicolumn{1}{c}{\bf Video} & \multicolumn{1}{c}{\bf Text} &
         \multicolumn{1}{c}{\bf 1-shot} & \multicolumn{1}{c}{\bf 5-shot} \\
    \midrule
    TSN++ & Inductive & \xmark & \xmark & $35.72 \pm 1.37$ &  $49.4 \pm 1.24$
    \\
    T-TSN+CNAPS & Transd &\cmark & \xmark & 
    $42.20 \pm 1.43$ & $57.68 \pm 1.37$
    \\\midrule
    VNI & Inductive & \cmark & \xmark & 
	$33.53 \pm 1.07$ & $50.51  \pm 1.30$ 
    \\
    TNI & Inductive &\xmark & \cmark & 
    $34.27 \pm 1.24$ & $56.21 \pm 1.22$
    \\
    VNT & Transd. & \cmark & \xmark & $41.5\pm 1.53$ &  $56.60 \pm 1.23$
    \\
    TNT & Transd. & \xmark & \cmark & $\textbf{47.98} \pm 1.41$ & $\textbf{60.18} \pm 1.21$
    \\
    \bottomrule
    \end{tabular}
} & \bmvaHangBox{
\includegraphics[scale=0.45]{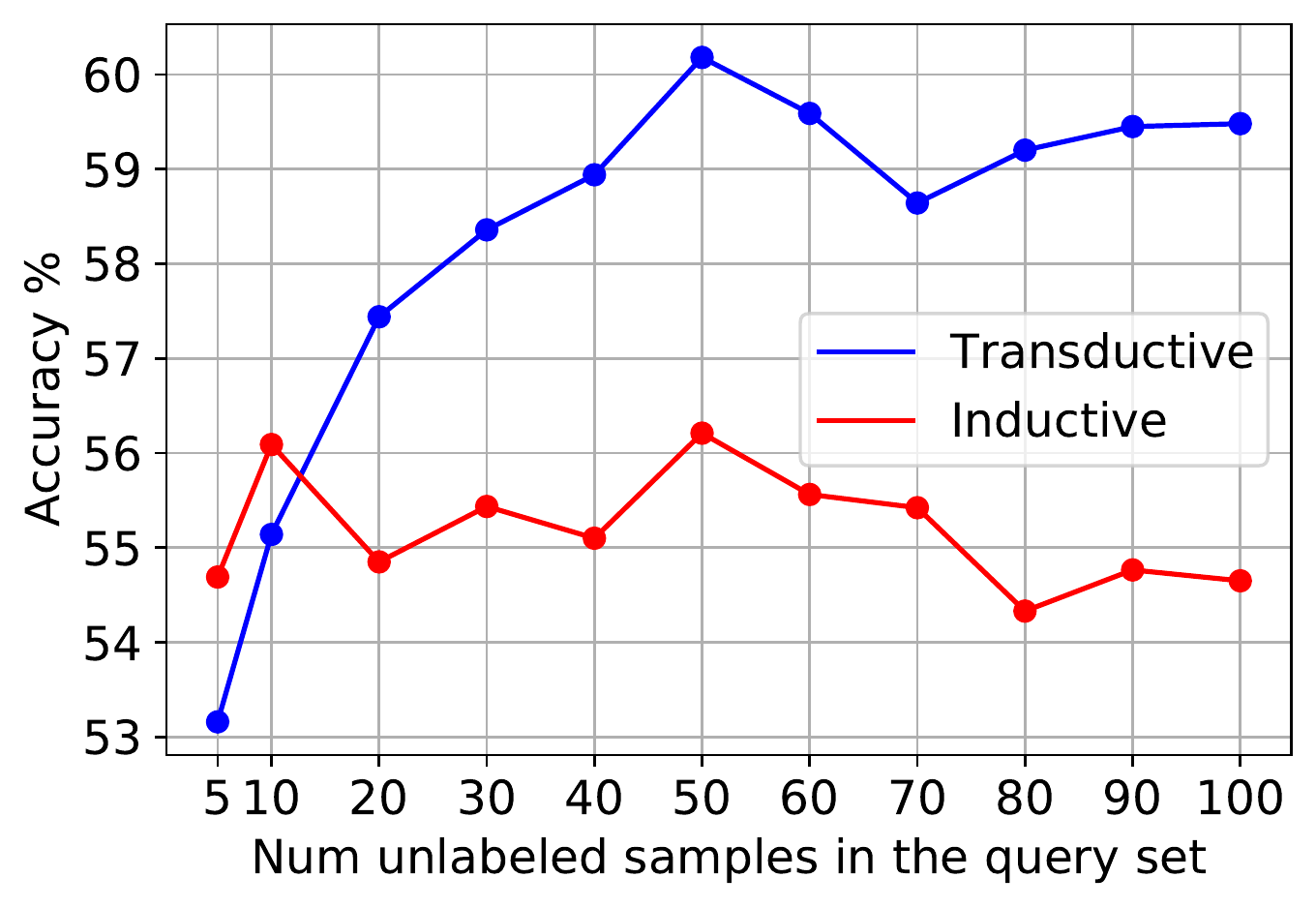}
}
\end{tabular}
}
\caption{(Left) \textbf{Ablation study results in the validation set.} We consider: a Video-conditioned Networks with Inductive (VNI) and Transductive (VNT) inference, Text-conditioned Networks with Inductive (TNI) and Transductive (TNT) inference, the TSN++ baseline, and the T-TSN+CNAPS baseline based on \cite{BateniSimpleCNAPS, liu2020prototype}.} 
\label{tab:ablation_studies}
\vspace{-1mm}
\captionof{figure}{
(Right) \textbf{Sensitivity analysis to the number of query set samples.} Model performance in the 5-way, 5-shot task for different B size.
}
\vspace{-4mm}
\label{fig:ablation_studies}
\end{table}

\noindent\textbf{Ablation Study.} We analyze the impact of our proposed text encoder and transductive inference approach. To this end, we train and evaluate our model with two different task encoders (text and video) and two inference approaches (Inductive and Transductive). Specifically, we consider video-conditioned networks with inductive ({\bf VNI}) and transductive ({\bf VNT}) inference; and text-conditioned networks with inductive ({\bf TNI}) and transductive ({\bf TNT}) inference. The video-based conditioner module consists of a C3D model with a Conv-4-64 backbone to extract significant spatio-temporal information at low computational cost \cite{zhang2020fewshot}, see Fig.~\ref{fig:encoders_dynaMod}-b. Moreover, we consider the TSN++ \cite{Cao_2020_TAM} as our primary baseline, which is purely based on \cite{TSN} and inductive prototypical learning. For a fair comparison, we also implement T-TSN+CNAPS, a variant of TSN++ with modulation \cite{BateniSimpleCNAPS} and transductive inference based on label propagation \cite{liu2020prototype}. Table \ref{tab:ablation_studies} shows our results. All the results are computed in the validation set using the same visual backbone model (ResNet-34).

It is important to note that feature encoder adaptation generates a substantial increase in model performance with regard to the TSN++ baseline across all evaluation modalities. This is generally true for both the VNI and TNI models. Crucially, the TNI model yields a performance gain of 6\% and 1\% in  5-shot and 1-shot tasks on the SS-100 over VNI, respectively. Likewise, there is a further performance increase, especially in the 1-shot task when the dynamic module is included to perform a transductive inference (TNT and VNT). Interestingly, the 1-shot task is the most data-deprived testing set-up, which speaks positively about the effectiveness of our transductive model. It should be noted that our TNT model outperforms the T-TSN+CNAPS and VNT models by a large margin on both 5-shot and 1-shot tasks. This proves the relevance of using textual descriptions to modulate or contextualize the video feature encoder and improve the class prototypes in a transductive approach.

\noindent\textbf{Effect of Query Set Size.}
We assess the TNT model sensitivity to the number of instances in the query set. This study can be observed in Fig. \ref{fig:ablation_studies}. 
We evaluate our model trained on SS-100 in the 5-way, 5-shot task with $B=50$, increasing the value of $B$ from 5 to 100.
Model performance increases until the number of query samples $B=50$ after which it remains almost constant. We hypothesize that this is due to a saturation point on the amount of extra information that can be extracted from query samples. 

We also conduct qualitative evaluations to demonstrate how our model works and the relevance of using textual descriptions to modulate the visual feature encoder and perform a transductive inference. They are shown in the supplementary material.

\section{Conclusions}
In this paper, we propose the Text-Conditioned Network with Transductive Inference (TNT), a novel few-shot model that leverages the fine-grained textual descriptions of the support instances to improve video understanding under a low-data regime. 
Unlike previous works, TNT uses text representations from a pre-trained language model to adapt and contextualize the feature encoder to each FSL task and improve class prototypes in a transductive setting.
Our experiments show that our model outperforms a wide range of state-of-the-art models in four challenging datasets.
Furthermore, our ablation study shows that the dynamic prototype module plays an important role in improving the 1-shot task. As an important finding, we verify that textual conditioning provides a more helpful signal than video-based conditioning to enhance the video feature encoder.
\section*{Acknowledgements}
This work was supported in part by the Millennium Institute for Foundational Research on Data (IMFD).

\bibliography{egbib}
\end{document}


\maketitle
\renewcommand\thefootnote{$\dagger$}
\footnotetext{Equal advising.}

\renewcommand\thefootnote{$\ast$}
\footnotetext{Current affiliation: Facebook AI, London, UK.}

In the main paper, we introduce a novel Few-Shot Learning (FSL) model for video action classification: \textbf{T}ext-Conditioned \textbf{N}etworks with \textbf{T}ransductive inference (TNT). This model leverages the semantic information in the textual descriptions of support instances as a privileged source of information to improve class discrimination in a scarce data regime. Specifically, we leverage the textual descriptions to modulate the visual encoder and compute transductive inference, augmenting the support instances with those unlabeled through an attention-based and multimodal approach. Overall, the integration of these abilities allows our model to adapt to the FSL tasks quickly. This supplementary material contains the following: \textbf{(1)} An overview of the main attributes of the four challenging video datasets used to evaluate our model; \textbf{(2)} Qualitative evaluations to demonstrate the relevance of using textual descriptions to modulate our network. Moreover, this document is accompanied by a video that presents the principal motivation and contributions of our approach.
\vspace{-3mm}

\section{Datasets}
\begin{figure}
\centering
\begin{tabular}{cc}
\resizebox{6cm}{!}{
\bmvaHangBox{
\includegraphics[scale=1.0]{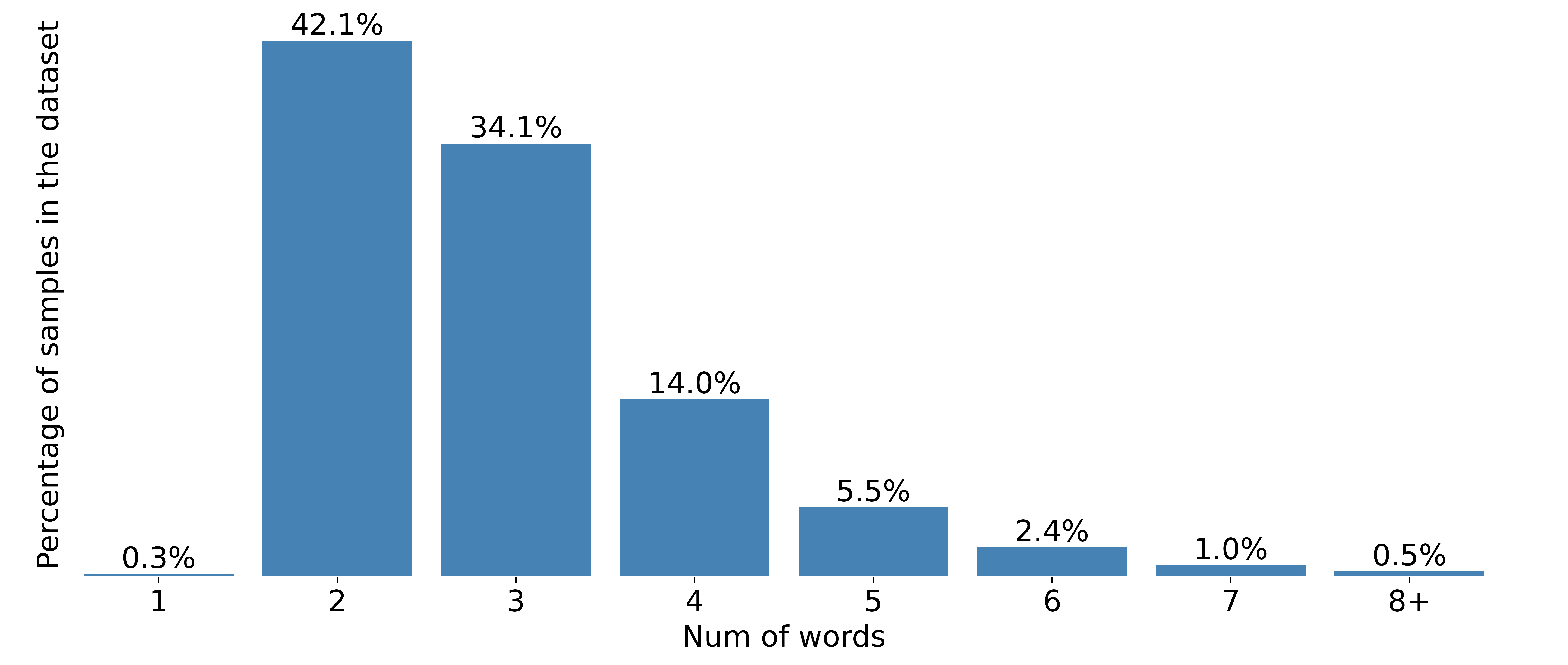}}
} & \resizebox{6cm}{!}{
\bmvaHangBox{
\includegraphics[scale=1.0]{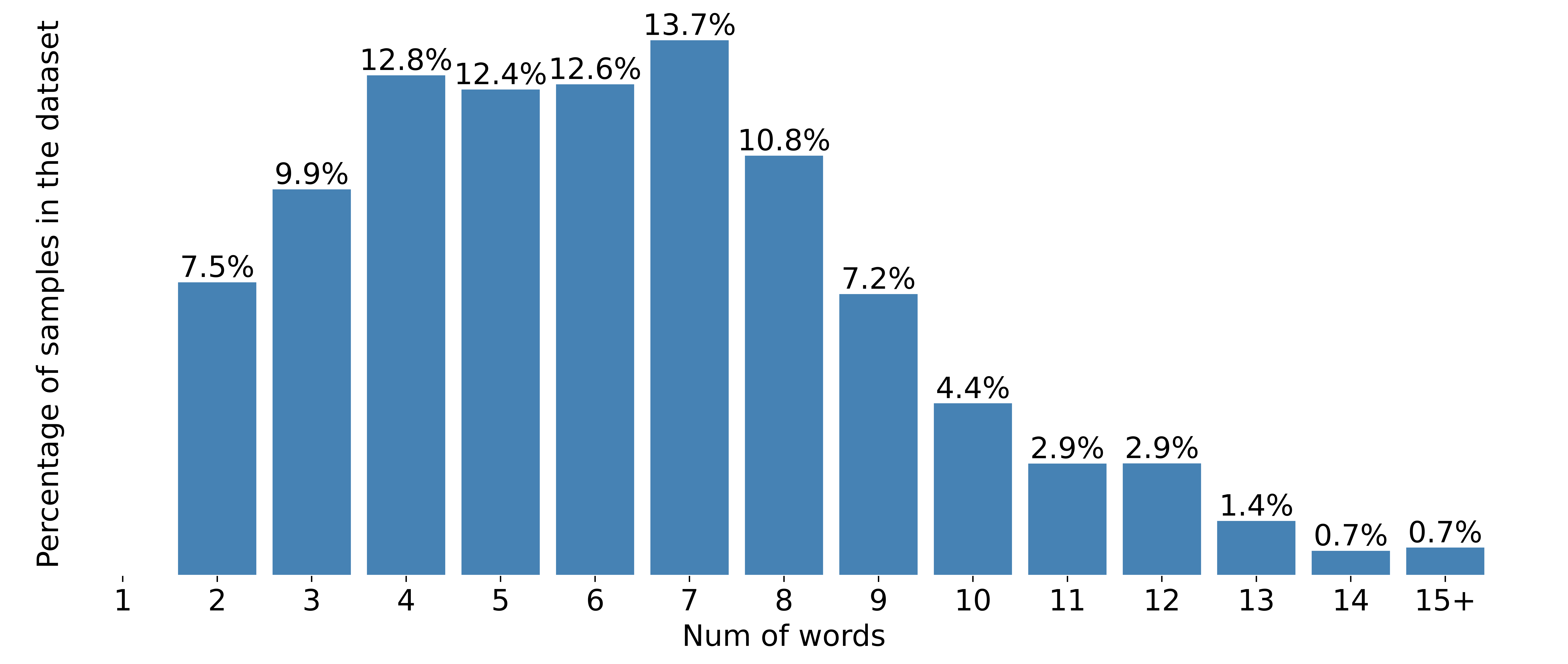}
}}\\(a) & (b)\\\resizebox{6cm}{!}{
\bmvaHangBox{
\includegraphics[scale=1.0]{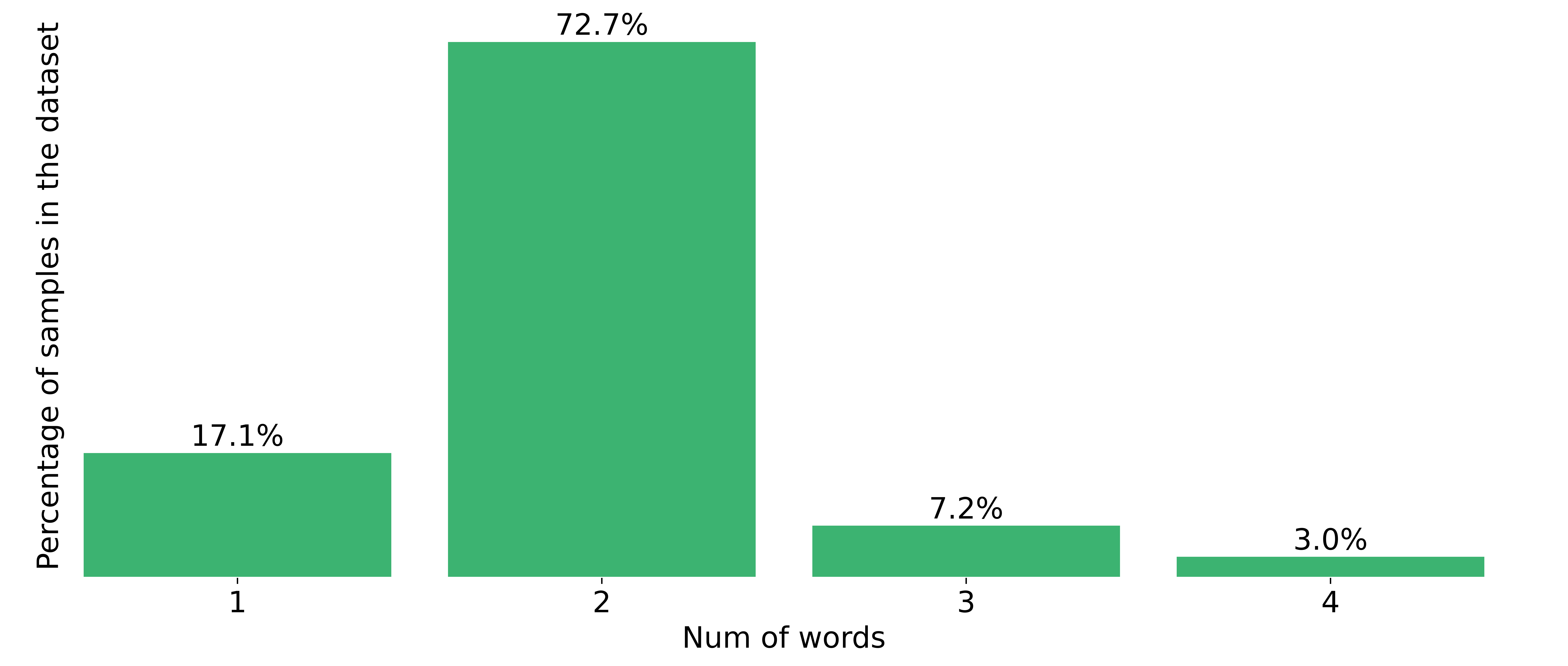}}
} & \resizebox{6cm}{!}{
\bmvaHangBox{
\includegraphics[scale=1.0]{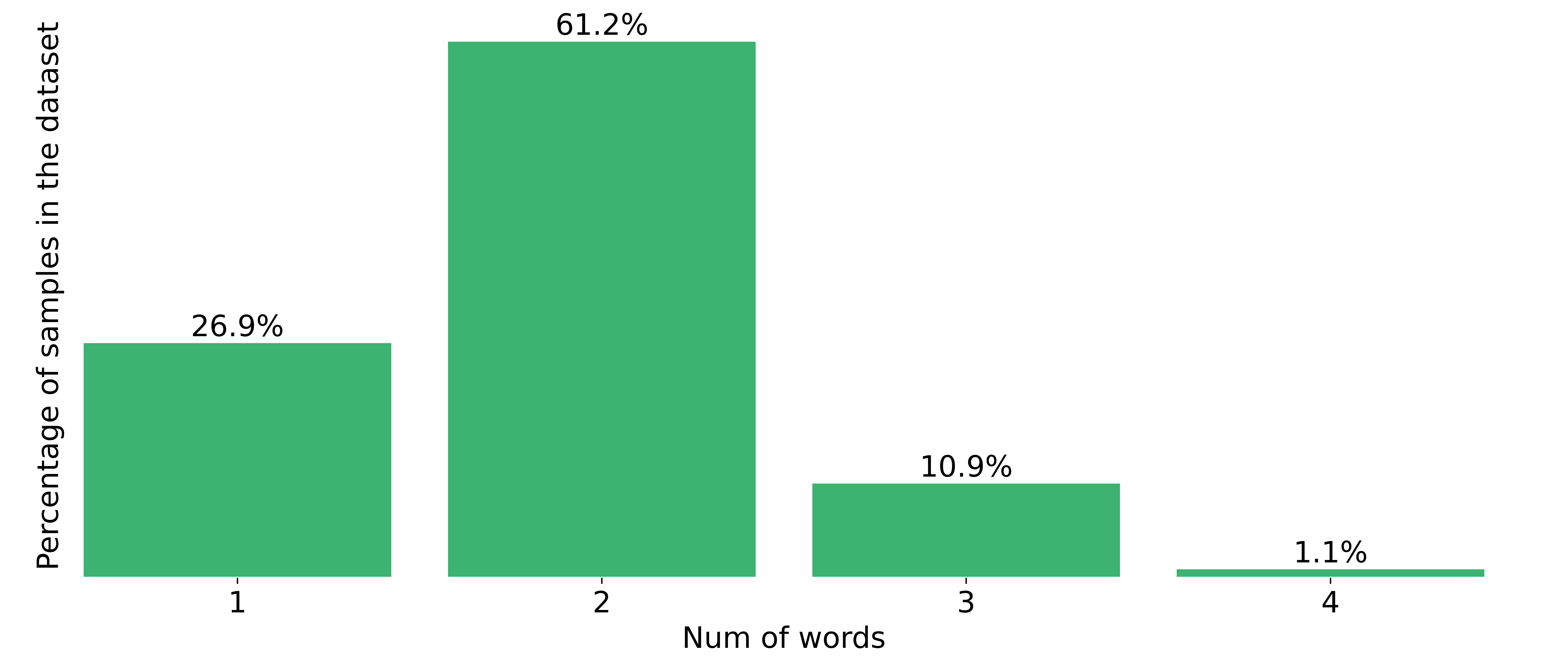}
}}\\(c) & (d)
\end{tabular}
\caption{
\textbf{Length of action descriptions.} Distribution of the number of words per instance description in both families of datasets. (Blue figures) Those with rich textual descriptions: (a) EK-92 and (b) SS-100~\cite{Cao_2020_TAM}. (Green figures) Those with short class-level descriptions: (c) Kinetics-100~\cite{compoundMem} and (d) MetaUCF-101~\cite{mishra2018generative}.
}
\label{fig:stads_datasets}
\end{figure}

\vspace{-2mm}
\begin{table}[t]
\centering 
\resizebox{8.3cm}{!}{%
    \begin{tabular}{cccccccc} 
    \toprule
    \multicolumn{1}{c}{\multirow{2}{*}{\bf Dataset}} &\multicolumn{3}{c}{\bf Num Classes} & \multicolumn{3}{c}{\bf Num Instances} & \multicolumn{1}{c}{\multirow{2}{*}{\begin{tabular}[c]{@{}c@{}}{\bf Rich Textual} \\ {\bf Descriptions} \end{tabular}}} \\
    \cmidrule{2-7}
         & Train & Val & Test & Train & Val & Test & \\
    \midrule
    EK-92 & 58 & 11 & 23 & 49621 & 8352 & 18370 & \cmark  
    \\
    SS-100 & 64 & 12 & 24 & 67013 & 1926 & 2857 & \cmark
    \\
    MetaUCF-101 & 70 & 10 & 21 & 9154 & 1421 & 2745 & \xmark
    \\
    Kinetics-100 & 64 & 12 & 24 & 6400 & 1200 & 2400 & \xmark
    \\
    \bottomrule
    \end{tabular}}
\caption{
\textbf{Statistics} from the datasets for FSL: SS-100~\cite{Cao_2020_TAM}, MetaUCF-101~\cite{mishra2018generative}, Kinetics-100~\cite{compoundMem} and the introduced EK-92.
}
\label{tab:datasets}
\end{table}
We evaluate our model on four challenging video action FSL benchmarks, those that contain rich textual descriptions (such as EK-92 and SS-100~\cite{Cao_2020_TAM}), and short class-level descriptions (such as MetaUCF-101~\cite{mishra2018generative} and Kinetics-100~\cite{compoundMem}) . Interestingly, in EK-92 and SS-100~\cite{Cao_2020_TAM}, the descriptions are specific per each video instance and can have more than 8 and 15 words, as is shown in Fig.~\ref{fig:stads_datasets}-a and Fig.~\ref{fig:stads_datasets}-b, respectively. Thus, its descriptions are detailed and very correlated with the actions and objects in the videos. Conversely, in Kinetics-100~\cite{compoundMem} and MetaUCF-101~\cite{mishra2018generative}, the descriptions correspond to the class label. Therefore, they are the same for the videos that belong to the same class. Fig.~\ref{fig:stads_datasets}-c and Fig.~\ref{fig:stads_datasets}-d show that Kinetics-100~\cite{compoundMem} and MetaUCF-101~\cite{mishra2018generative}, respectively, have short descriptions with no more than 2 words for most of them. Notably, our model achieves outstanding results on the four benchmarks, even on those with short class-level descriptions. Table~\ref{tab:datasets} summarizes the main attributes of these benchmarks, which are presented in the Dataset section of the main paper.

\vspace{-3mm}

\section{Qualitative results}
\begin{figure}
\centering
\begin{tabular}{cc}
\resizebox{5.5cm}{!}{
\bmvaHangBox{
\includegraphics[scale=0.5]{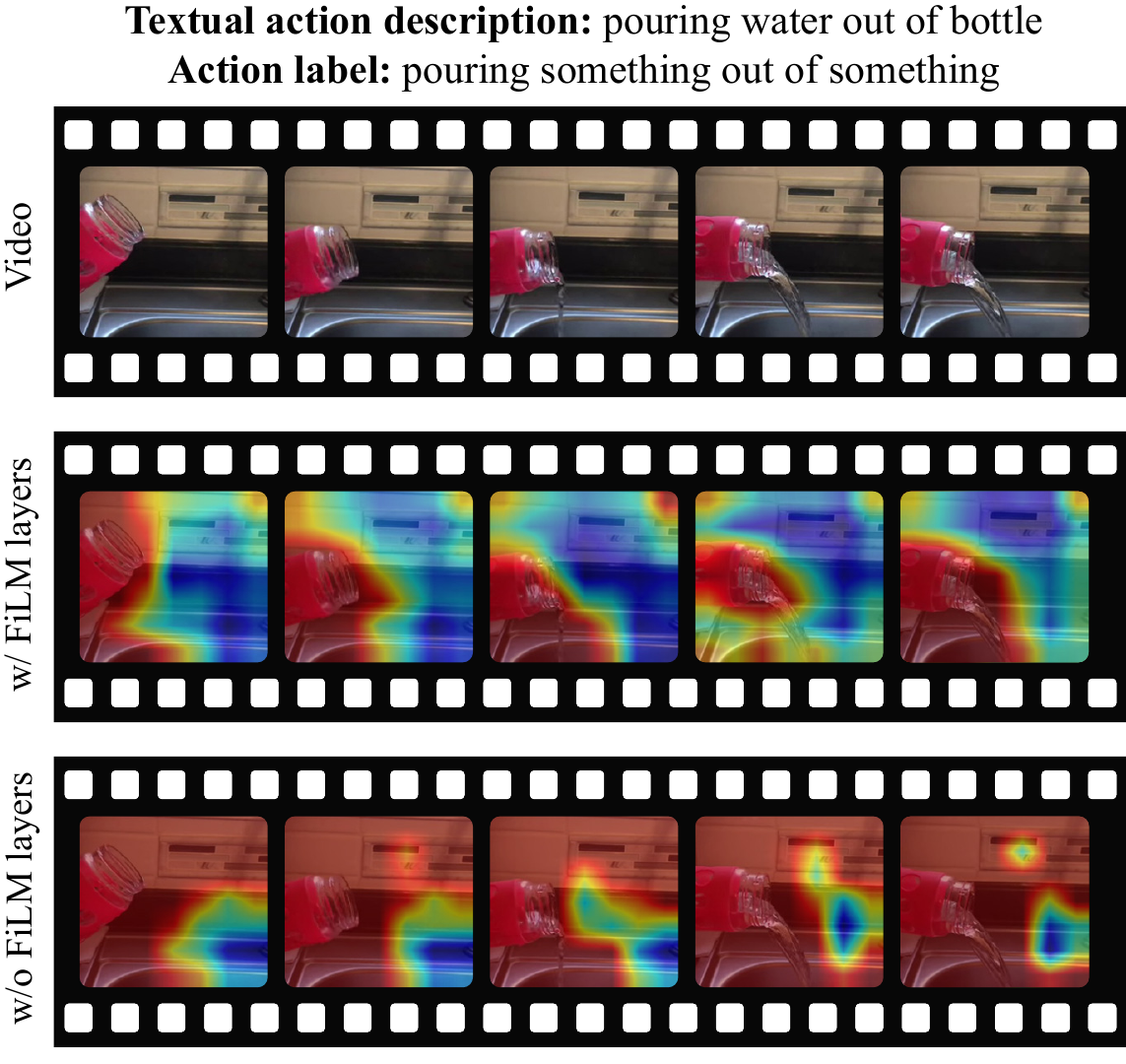}
}} & \resizebox{5.5cm}{!}{
\bmvaHangBox{
\includegraphics[scale=0.4]{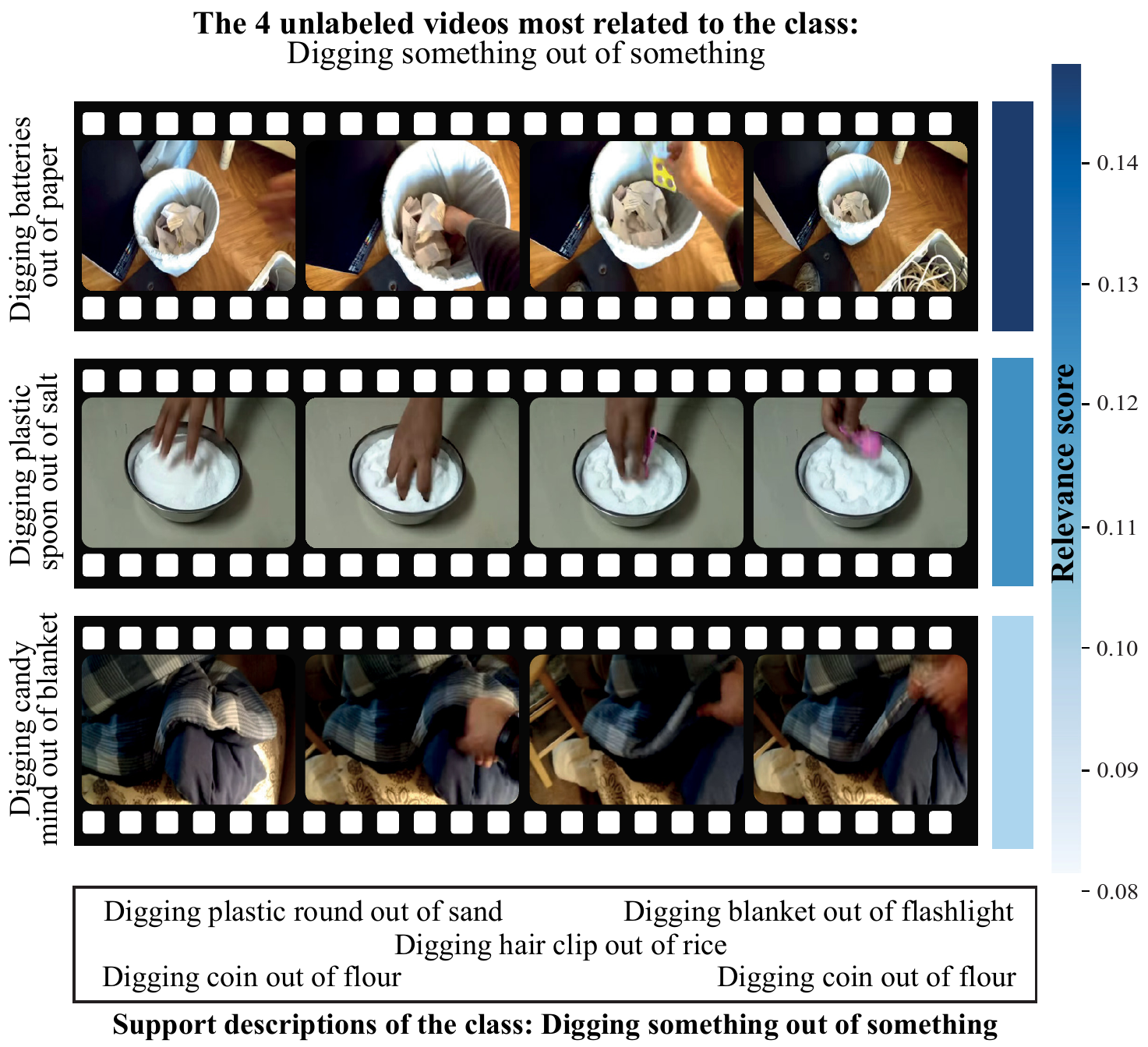}
}}\\(a) & (b)
\end{tabular}
\caption{
\textbf{Visualizations.} (a) We take our model pre-trained on SS-100 in the 5-shot and 5-way task and analyze the changes of its CAMs with and without using the FiLM layers before the last average pooling layer of the ResNet. (b) Top-3 relevant unlabeled samples for the class in a 5-way 5-shot task. The darkest color indicates the most important sample.
}
\label{fig:visulizations}
\end{figure}

\vspace{-2mm}
\noindent\textbf{Visualization of Class Activation Maps.}
An important aspect of our proposed method is its ability to adapt the feature backbone to a particular task. We are interested in verifying the effect of our proposed text-conditioned module in video samples. To study the influence of FiLM layers to modulate the TSN network using semantic information from textual action descriptions, we analyze the class activation mapping (CAM) for the cases with and without employing the FiLM layers. We use our TNT model trained on SS-100 for the 5-shot 5-way task. Fig.~\ref{fig:visulizations}-a shows the visualization obtained for one test sample after applying our TNT model with and without FiLM layers following the Grad-CAM \cite{grad_cam} approach. 

As it can be seen, the TNT model manages to exploit relevant cues from the support textual descriptions. In this example, our FiLM-ed model activations are located in appropriate regions for the action ``Pouring water out of bottle''. Of particular interest, it seems that the effect of our text-conditioned features is to allow the underlying model to focus on the scene elements that matter the most for the particular query sample. This is, in this sample, the bottle and the pouring water. Meanwhile, the model with no FiLM layers tends to attend irrelevant regions. This arguably has effects in a diminished capacity for fine-grained classification.
In general, we found that the conditioning process, based on the textual action descriptions, enables the network to obtain better video features for specific tasks.


\noindent\textbf{Dynamic Module Inspection.}
To assess the transductive capabilities of our model, we closely study the proposed dynamic module.  We exploit our SS-100-trained TNT model in the 5-way 5-shot setting. For this experiment, we directly observe the values of the relevance weight matrix $\mathbf{W}$, which can be easily interpreted. 
Indeed, they encode how relevant is each one of the query set videos to the semantic class embedding obtained by textual descriptions of the support set $\textbf{E}_{class}^{\mathcal{T}}$.
Fig.~\ref{fig:visulizations}-b shows the three most related videos from the query set $\queryset$ to the ``digging something out of something'' class. The text-driven class representation is obtained by averaging the semantic embedding of all sampled descriptions for this class in the support set. These descriptions are shown at the bottom of the figure.
Furthermore, in this figure, we present a heatmap in which darker color means greater relevance of the query video sample to the actual support set description.
Interestingly, the top three most related videos from the query set are the ones that also belong to the class ``digging something out of something''.
This example demonstrates that our dynamic module can leverage the semantic information of textual action descriptions to augment the support set samples with unlabeled samples from the query set. 
This behavior of our model is essential to improve class prototypes and alleviate the problem of low training data. 

\vspace{-3mm}




\bibliography{egbib}